\pdfoutput=1
\documentclass[10pt, conference]{IEEEtran}

\usepackage{cite}
\ifCLASSINFOpdf
\usepackage{graphicx}
\else
\fi
\usepackage[cmex10]{amsmath}
\usepackage{array}
\usepackage{url}
\usepackage{multirow}
\usepackage{float}

\begin{document}

\title{Supervising Unsupervised Learning with Evolutionary Algorithm in Deep Neural Network}

\author{\IEEEauthorblockN{Takeshi Inagaki}
\IEEEauthorblockA{IBM Japan, Tokyo}
}
\maketitle

\begin{abstract}
A method to control results of gradient descent unsupervised learning in a deep neural network by using evolutionary algorithm is proposed. To process crossover of unsupervisedly trained models, the algorithm evaluates pointwise fitness of individual nodes in neural network. Labeled training data is randomly sampled and breeding process selects nodes by calculating degree of their consistency on different sets of sampled data. This method supervises unsupervised training by evolutionary process. We also introduce modified Restricted Boltzmann Machine which contains repulsive force among nodes in a neural network and it contributes to isolate network nodes each other to avoid accidental degeneration of nodes by evolutionary process. These new methods are applied to document classification problem and it results better accuracy than a traditional fully supervised classifier implemented with linear regression algorithm.
\end{abstract}


\IEEEpeerreviewmaketitle

\section{Introduction}

Deep Learning applying unsupervised learning to lower layers (near to input layer) and training higher layers (near to output layer) with labeled data captures wide variety of data characteristics from unlabeled data and supervised learning layer can recognize abstracted features of target data \cite{Hinton}. Data in some categories, such as images or sound or other data detected by sensors, are governed by laws of physics behind them and it is expected that unsupervised learning automatically detects patterns in data caused by these laws. These patterns are regarded as features in data used for supervised leaning of higher layer. However, if this method is applied to more conceptual level of problem such as classification of text documents described by natural human language, features extracted by unsupervised layers are not always relevant for labeling of data \cite{McAfee}. To understand the reason, we need to know the role of unsupervised learning in lower layers. Unsupervised learning layers are designed to have large number of input nodes and less number of output nodes. It results to reduce number of dimensions of parameter space representing input data. In case of image recognition, not all possible alignments of bit pixels occur in photo images. Only limited number of patterns of alignments happen in the real world governed by laws of physics, like face of cat, shape of woods in forest, tall building and so on. By applying unsupervised learning to neural network, it is expected to reduce bit pixels parameters to limited number of image patterns. These patterns with reduced number of degree of freedom are regarded as representations of concepts human recognizes. By referring these abstracted features, supervised learning performed on higher layer of network does not need large amount of labeled data to cover all pixel alignment patterns but just needs for limited number of patterns of pixels to be labeled. When the same method is applied to text documents, unsupervised learning layers are expected to form clusters of words or phrases representing abstracted features. A problem arisen for classification of text documents is that concepts in human writings are not always concrete but sometimes rather abstract and there is ambiguity in its nature. For example, a description by text may allow multiple interpretations depending on interest of individual person who reads it. At attempt of classification of documents, one text includes multiple concepts and some of them are relevant for specific classification but are not useful for other classification context. Concepts automatically detected by unsupervised learning contain these concepts unnecessary for a specific classification program and they may cause unexpected results in classification. To avoid this, we develop new evolutionary algorithm which picks nodes in neural networks relevant for a specific classification problem to be addressed and generates a child neural network from them. This new method results better classification accuracy than traditional classifier using the linear regression algorithm. It works efficiently especially when there is large amount of unlabeled data but is small number of labeled data.

\section{ Evolutionary Algorithm}

Idea of Evolutionary Algorithm for neural network (called Neuroevolution) has rather long history \cite{ Yao} and is still an active area for study \cite{Such}. It was inspired by evolution of life and is designed to adapt neural network to external environment. It evaluates fitness of parents to environment and breeds children from better fitting individuals. Basic technique employed are crossover and mutation. This approach is regarded as a method to find a solution of optimization problem by random sampling (or namely global search), and it is an alternative to traditional gradient descent method which assumes analytic property (differentiability) of a loss function in the parameter space and the learning process is traveling on a trajectory to reach to a minimum point of the function. A new method introduced in this paper applies evolutional steps of evolutionary algorithm to overcome the issue discussed in previous section. Geometrically, there is no distinction between relevant features and irrelevant features extracted by unsupervised learning and they are detected by gradient descent equally. In the crossover process of evolutionary algorithm, selection of features is performed by sampling method. This can be done by introducing a metric to measure relevancy of features. This new method turns out to be effective for cases where gradient descent partially works but it can't fit a solution perfectly due to the global ambiguity of underlying unsupervised learning results by which not all of geometric minimums are relevant for a solution. Evolutionary process selects relevant one and removes irrelevant one.

\section{ Supervising Evolutionary Algorithm}

New algorithm proposed here is hybrid of evolutionary algorithm and gradient descent method. We use a simple model with three layers of nodes. Nodes in a layer are connected to nodes to other layers, upper or lower layer. Nodes in the middle layer (hidden layer) correspond to concepts in data and each of them is connected to nodes in the input layer and the output layer. In case of document classification problem, nodes in the input layer are corresponding to words in a document and nodes in the output layer correspond to categories of classification. In more general cases, input nodes are corresponding to features extracted from target data e.g. shapes of images and so on. Connections among nodes are represented by weight factors (denoted as $w_{i,j}$ below). The hidden layer is trained by unsupervised training algorithm (autoencoder is used in this paper) to form concepts in data, for example, clusters of words in the context of document classification. To define statistical mechanics model for this hidden layer, the Restricted Boltzmann Machine described by action (1) is employed 

\begin{equation}
S(x,y) = - \sum_{i,j} w_{i,j} y_i x_j + \sum_i b_i y_i + \sum_j c_j x_j
\end{equation}

where $y_i$ represents concepts and $x_j$ represents input features such as appearance of words in a document. With this action, a probability distribution function of concepts for input features is given by (2).

\begin{equation}
P(y | x) = e^{-S(x,y)}
\end{equation}

The output layer on top of the hidden layer is trained with labeled data to associate concepts in hidden layer to categories of classification. This layer is trained by the backwards propagation to correct incoincidence of prediction with label data. 

Idea of evolutionary algorithm is to execute these two steps of training in iterations and every iteration uses bred artifacts of previous iteration as a set of initial values of model parameters for next training. During these continuous steps, it generates a child model from multiple parent models. A child is built up from hidden layer nodes picked from parents. These nodes are selected by evaluating relevancy for given classification problem. In that way, later generation of a model contains concepts relevant for a classification problem and will have better accuracy for that problem.

This process is described in below:
\begin{enumerate}
\item At first a few iterations, models are populated with initial seed parameters randomly generated. First, train the hidden layer with unlabeled data by unsupervised training, then the train output layer with labeled data. These learning processes employ gradient descent. Resulted models are put in a pool of models.
\item Then after, examine models in the model pool with labeled test data and select two or more models with good precision score.
\item In the hidden layer of selected models, pick hidden nodes (concepts) consistently contributing to classification. A function to measure consistency is defined later in this paper.
\item Breed a new hidden layer combined with nodes picked in previous step, create a new output layer with random parameters.
\item Train a new hidden layer with unlabeled data and train a new output layer with labeled data by gradient descent again. Then this new model is put in the model pool.
\item Repeat process from 2) to 5) until model growth of model accuracy saturated.
\end{enumerate}

More details of steps 3) and 4) are as the following. Pick $M$ models in the model pool and perform supervised training of the output layer. At that moment, we divide labeled documents for training in $N$ sets in random but to include all categories equally. With these training sets train $N$ of individual output layers on top of a same hidden layer. Do this operation for all $M$ hidden layers. This process generates $M \times N$ models. Evolutionary process is picking relevant nodes in the hidden layer which contribute to classification consistently for all $N$ models trained on top of that hidden layer. Doing same for all $M$ of independent hidden layers with $N$ output layers for each, we obtain a set of nodes of hidden layers picked from $M$. To evaluate relevancy of each node in the hidden layer, a score value calculated by the function (3) is used

\begin{equation}
s_j =\sum_{n' \ne n} \sum_i w_{i, j}^{n} w_{i, j}^{n'}
\end{equation}

where $w_{i,j}^{n}$ is weight parameters of the output layer and index $j$ stands for the hidden layer node (concept) and $i$ for the output layer nodes (category) and $n$ is index of $N$ training sets of the output layer. This regards $ w_{i,j}^{n}$ as a vector $|w_{j}^{n}>$ spanning in dimensions of categories. The score (3) is sum of inner products of two vector $|w_{j}^{n}>$ and $|w_{j}^{n'}>$ for all combination of $n$, $n'$ but $n \neq n'$. Here inner product is used to measure coincidence of concepts $j$ in document set $n$ and $n’$ over all categories. To understand implication of this, we need to know cause of misdetection of categories. Concepts detected by the autoencoder are based on deviation of distribution of words on each document. This may or may not be related to concepts relevant for categories. If we do supervised training on top of these concepts, some of irrelevant concepts may correlate to categories by accident. However, this is accidental occurrence only on a set of training documents used and it may not happen for other sets of documents. Because there is no relation between sampling based dividing of training documents and categories of documents with labels, if we do sampling multiple times, effect of labeled categories will be persistent but effects of accidental co-occurrence in sampled documents will disappear. For example, let us consider the case both of relevant and irrelevant concepts appear on a set of documents of a category with probability 50\%. If we take another document set of the same category, relevant concepts may appear in it again consistently with probability 50\% but same irrelevant one may with 25\% just by double of accidents. With this observation, we can conclude that if a concept is relevant for categories, it will be correlated to them in different sample sets of documents. In a simplest case of $N=2$, labeled documents are divided in two sets, even and odd, and both sets includes all categories of documents equally. Equation (3) becomes $s_j = \sum_i w_{i, j}^{even} w_{i, j}^{odd}$. This measures how a concept $j$ correlated on all categories in even and odd sets. If the concept is relevant, this inner product becomes large. By applying this measurement in the iteration process described above, nodes in the hidden layer continuously becomes more relevant for categories of classification.

\section{Implementation}
We apply Restricted Boltzmann Machine in hidden layer and train it with Denoising Autoencoders \cite{Vincent}. 
By performing autoencoding and decoding steps and adjust parameters $w$, $b$, $c$ by the gradient descent to minimize the cross entropy loss function (4)

\begin{equation}
L = \sum_{j} \{x_j ln \hat{x}_j + (1-x_j) ln (1-\hat{x}_j)\}
\end{equation}

where, $\hat{x}$ is reconstructed visible parameter calculated from equations (5) and (6).

\begin{eqnarray}
E_i & = & \sum_{j} w_{i,j} x_j + b_i \nonumber \\
E_j & = & \sum_{i} w_{i,j} \hat{y}_i + c_j
\end{eqnarray}

\begin{eqnarray}
\hat{x}_j & = & \frac{1}{1 + e^{-E_j}} \nonumber \\
\hat{y}_i & = & \frac{1}{1 + e^{-E_i}}
\end{eqnarray}

One potential problem with the algorithm described in previous section is condensation of multiple nodes in the hidden layer on a single concept. That means multiple nodes in the hidden layer represents a same concept redundantly and do not contribute for classification individually. To avoid that, we implemented repulsive force among nodes in Restricted Boltzmann Machine. It can be achieved by modifying energy of the Boltzmann Machine states as (7)

\begin{eqnarray}
\tilde{E}_i & = & \sum_{j} (w_{i,j} -\alpha \sum_{i' \neq i} w_{i', j}) x_j + b_i \nonumber \\
E_j & = & \sum_{i} w_{i,j} \hat{y}_i + c_j
\end{eqnarray}

where $\alpha$ is a small constant proportional to $O(\frac{1}{\sqrt{C}})$ and $C$ is number nodes of output layer (number of categories). This modification changes the stochastic gradient descent as (8)-(10). With this, influence of input nodes (words) commonly referred in many nodes in the hidden layer are suppressed.

\begin{eqnarray}
\frac{\partial L}{\partial w_{i,j}} & = & \sum_{i'} \frac{\partial L}{\partial \tilde{E}_{i'}} \frac{\partial \tilde{E}_{i'}}{\partial w_{i,j}} + \sum_{j'} \frac{\partial L}{\partial E_{j'}} \frac{\partial E_{j'}}{\partial w_{i,j}} \nonumber \\
& = & \sum_{i'} \sum_{j'} \{ w_{i',j'} (x_{j'} - \hat{x}_{j'})\} y_{i'} (1-y_{i'}) \nonumber \\
& & (\delta_{i,i'} - \alpha \sum_{i'' \neq i'} \delta_{i, i''}) x_j + (x_j -\hat{x}_j) y_i \nonumber \\
& = & [ \{ \sum_{j'} w_{i,j'} (x_{j'} - \hat{x}_{j'}) \} y_{i} (1-y_{i})\nonumber \\
& & - \alpha \sum_{i'' \neq i} \{ \sum_{j'} w_{i'',j'} (x_{j'} - \hat{x}_{j'}) \}y_{i''} (1-y_{i''}) ] x_j \nonumber \\
& & + (x_j -\hat{x}_j) y_i \nonumber \\
\end{eqnarray}

\begin{eqnarray}
\frac{\partial L}{\partial b_{i}} & = & \sum_{i'} \frac{\partial L}{\partial \tilde{E}_{i'}} \frac{\partial \tilde{E}_{i'}}{\partial b_{i}}\nonumber \\
& = & \{ \sum_{j'} w_{i,j'} (x_{j'} - \hat{x}_{j'}) \} y_{i} (1-y_{i}) \nonumber \\
& & - \alpha \sum_{i'' \neq i} \{ \sum_{j'} w_{i'',j'} (x_{j'} - \hat{x}_{j'}) \}y_{i''} (1-y_{i''}) \nonumber \\
\end{eqnarray}

\begin{eqnarray}
\frac{\partial L}{\partial c_{j}} & = & \sum_{j'} \frac{\partial L}{\partial E_{j'}} \frac{\partial E_{j'}}{\partial c_{j}} \nonumber \\
& = & x_j -\hat{x}_j
\end{eqnarray}

To stabilize learning results, ensemble of the hidden layer is populated. The input layer is connected multiple hidden layers independently trained and evolved, the output layer is connected to all of nodes in this ensemble of the hidden layers. In stead of one big hidden layer with large number of concepts, a model has a set (ensemble) of independent hidden layers with smaller number of concepts and it contributes to improve accuracy in experiments.

\section{ Experimental results}

We classified 2,000 documents of 20 categories (100 documents in each). A hidden layer consists of 40 nodes and a model has 10 independent hidden layers for ensemble learning. To compare the result with a traditional method, linear regression algorithm is applied for same data. Both of linear regression model and this neural network model were trained with 200 labeled documents (10 documents in each). But in case of our new algorithm, 20,000 documents in the same corpus are used for unsupervised learning. Accuracy of linear regression model was 38.9\%. When Tf-Idf calculated from 20,000 documents was applied, the result of the linear regression algorithm was improved to 42.8\%. On the other hand, accuracy of this new model was 56.2\%. This accuracy largely varies with number and quality of documents used for unsupervised training. Fig. 1 shows increasing of accuracy by iteration for 2,000 documents were used for unsupervised training (46.1\%), 20,000 documents were used (56.2\%) and artificial documents generated from 2,000 documents by assembling words in same categories repeatedly in 20,000 documents (84.4\%). This artificial document set is regarded as a case where ideally large number of documents belonging to categories are used for unsupervised training.

\begin{figure}[H]
\begin{center}
\includegraphics[width=2.5in]{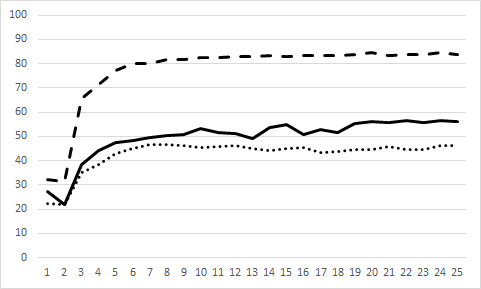}
\end{center}
\caption{\small Increasing of classification accuracy by iteration for different number of documents used for unsupervised training. Trained with 2,000 documents (dotted line 46.1\%), trained with 20,000 documents (solid line 56.2\%) and trained with artificially generated documents to simulate an ideally large number of documents (dashed line 84.4\%). Supervised training was performed with 200 documents (10 labeled documents for each of 20 categories) for all cases. Vertical axis is percentage of accuracy, horizontal line is number of iteration. Accuracy is improved by increase of number documents used for unsupervised training.}
\label{ResultGraph1}
\end{figure}

Fig. 2 indicates effect of repulsive force introduced by (7). When this effect is absent, increasing of accuracy was slower and it remained lower even after repeating iterations.

\begin{figure}[H]
\begin{center}
\includegraphics[width=2.5in]{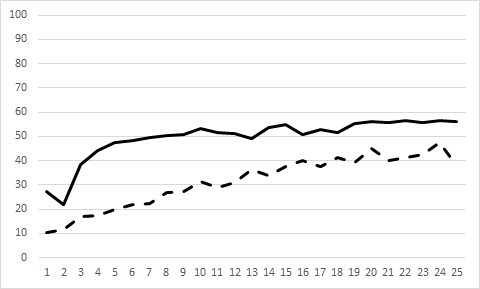}
\end{center}
\caption{\small Compare accuracy increasing by iteration for repulsive force applied case (solid line 56.2\% at max) and not applied case (dashed line 47.5\% at max). Vertical axis is percentage of accuracy, horizontal line is number of iteration. Without repulsive force, accuracy is lower and unstable.}
\label{ResultGraph2}
\end{figure}

Note that Tf-Idf, which was efficient for the linear regression model, did not improve accuracy if it is applied to new algorithm proposed in this paper. The reason is supposed to be that the same effect is already incorporated by repulsive force introduced by (7).

\section{ Conclusion}

New methods introduced in this paper is confirmed to work efficiently especially when only small number of labeled data is available but there is large amount of unlabeled data. This is important for practical use cases because labeled data is to be created by human and preparation of labeled data is effortful work. These methods are also applicable to classification of data other than text documents. For image recognition, if categories are directly corresponding to shapes of images, traditional deep learning works well. But if categories are more conceptual, the same problem as document classification appears in image recognition. Here extracted features of images by deep learning are corresponding to words and combination of them represents more complicated and non-trivial concepts. In such a case, method proposed in this paper helps to omit unnecessary features for categorization.

\bibliographystyle{unsrt}

\end{document}